\documentclass[letterpaper,twocolumn,10pt]{article}
\usepackage{usenix-2020-09}

\usepackage{tikz}
\usepackage{amsmath}

\usepackage{subfigure}
\usepackage{multirow}
\usepackage{booktabs}
\usepackage{blindtext}
\usepackage{makecell}
\usepackage{tabularx}
\usepackage{tikz}
\usepackage{amsmath}
\usepackage{amsthm}
\usepackage{mathtools}
\usepackage{multirow,multicol}
\usepackage{multicol}
\usepackage{ifthen}

\usepackage{amssymb}
\usepackage{optidef}
\usepackage{wrapfig}
\usepackage{graphicx}
\usepackage{subcaption}
\usepackage{float}
\usepackage[ruled,vlined]{algorithm2e}

\newcommand{\circled}[1]{%
  \tikz[baseline=(char.base)]{
    \node[shape=circle,draw,inner sep=0.5pt] (char) {#1};
  }%
}
\newcommand{\sysname}[0]{{ReFlow}}

\begin{document}

\date{}

\title{\Large \bf ReFlow: Replay-Driven Decoupling for Efficient RL Post-Training}

\author{
{\rm Yukun Tian\textsuperscript{1,3},
Junwei Su\textsuperscript{2},
Haofeng Xu\textsuperscript{1,3},
Lansong Diao\textsuperscript{3},
Zhengping Qian\textsuperscript{3},}\\
{\rm Weibo Zhao\textsuperscript{3},
Wenchao Zhou\textsuperscript{3},
Ruiting Zhou\textsuperscript{4},
Chuan Wu\textsuperscript{1}\thanks{\textsuperscript{}Corresponding author.}}\\[0.5em]
\textsuperscript{1}The University of Hong Kong
\qquad
\textsuperscript{2}University of Science and Technology of China\\
\textsuperscript{3}Alibaba Group
\qquad
\textsuperscript{4}Southeast University
}

\newcommand{\cwu}[1]{{\textcolor{red}{CWu: #1}}}
\newcommand{\js}[1]{{\textcolor{blue}{Js: #1}}}
\newcommand{\TODO}[1]{{\textcolor{gray}{TODO: #1}}}

\maketitle

\begin{abstract}
Reinforcement learning (RL) has been a critical paradigm for improving large language model (LLM) reasoning and agentic capabilities, but its efficiency remains a major bottleneck. 
Existing RL systems commonly rely on asynchronous execution to overlap rollout generation with policy optimization for enhanced efficiency. However, this approach faces inherent limitations: mild policy staleness offers only limited training efficiency gains, whereas aggressive asynchrony can degrade model quality and yields diminishing efficiency gains.
The dynamic nature of LLM RL and multi-turn agentic workloads further challenges RL system design, as resource demands shift over time, making static resource allocation and generation strategy inefficient. 

This paper presents \sysname{}, a system-algorithm co-designed RL framework for efficient LLM post-training. 
Its key insight is that, unlike existing RL frameworks which consume each trajectory only once, experience replay of recent trajectories enables effective sample reuse, thereby amortizing the high cost of rollout generation without incurring excessive policy staleness.
To fully exploit this opportunity, \sysname{} adopts a decoupled and adaptive architecture with three key designs: a dual replay data cycle supports efficient experience reuse while balancing training throughput and optimization stability; an elastic resource pool enables dynamic resource placement to accommodate evolving RL workloads; an adaptive scheduler coordinates generation strategies and resource allocation to improve resource utilization and boost generation throughput. 
Experiments across representative single-turn reasoning and multi-turn agentic RL tasks show that \sysname{} improves training throughput by up to $2.06\times$ over state-of-the-art synchronous and asynchronous RL systems, while maintaining comparable or better model performance.
\end{abstract}

\section{Introduction}\label{sec:intro}

\begin{figure}[!t]
    \centering
    \includegraphics[width=\linewidth]{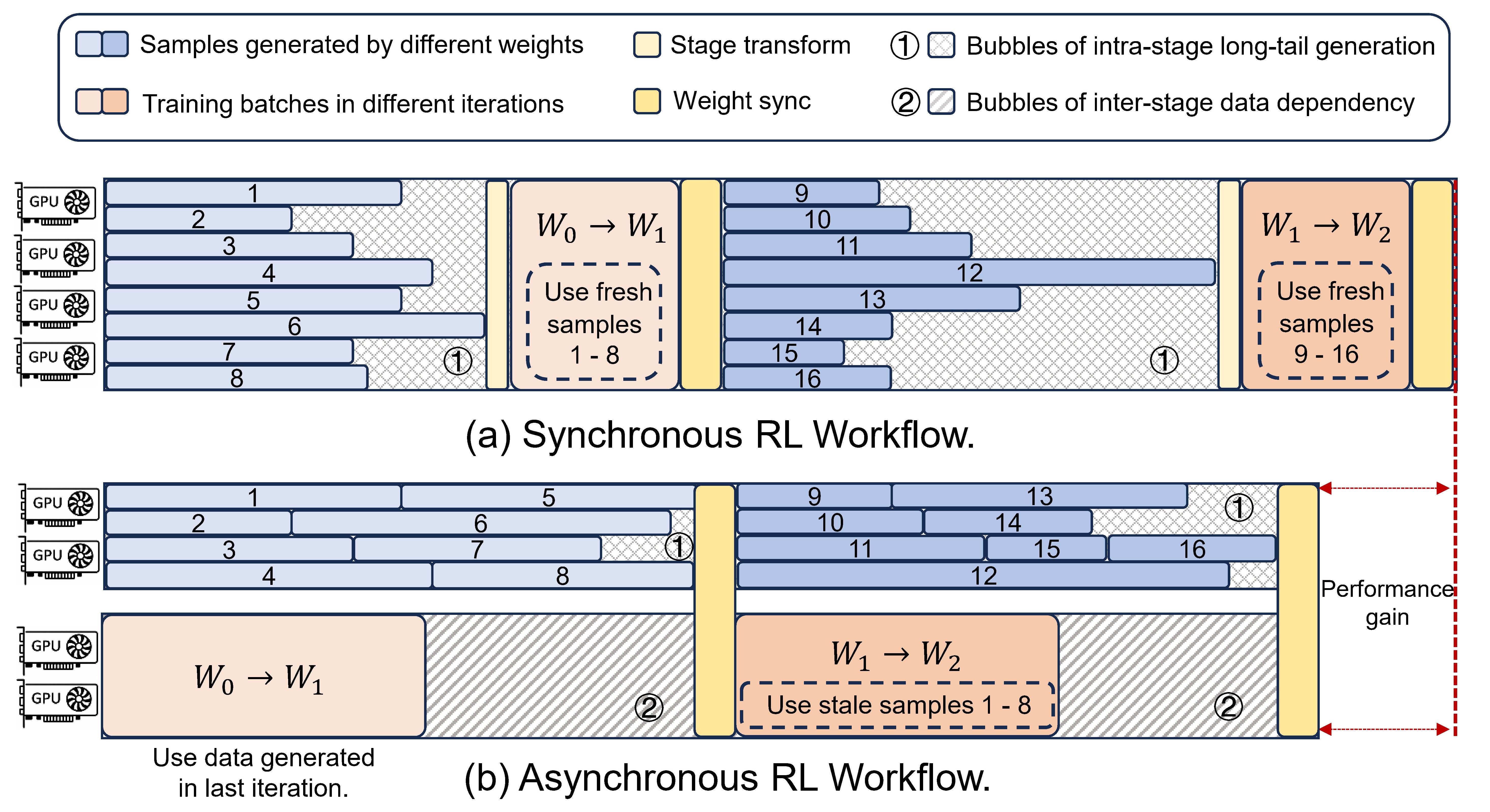}
    \caption{Typical LLM RL workflow: (a) synchronous RL; (b) asynchronous RL. $W$ denotes different versions of weights.
    }
    \label{fig:rlworkflow}
    \vspace{-4mm}
\end{figure}

Reinforcement learning (RL) has been widely adopted in post-training large language models (LLMs), enabling model alignment with human preferences~\cite{rlhf,rlhf1}, enhanced reasoning~\cite{deepseekr1,o1} and interactive agentic capabilities~\cite{glm5,deepseek3.2}. In RL post-training, the pretrained LLM iteratively generates responses to prompts, receives reward signals from human or automated feedback, and updates its policy to maximize expected return.
In practice, this workflow is typically decomposed into two primary stages—rollout generation and policy training—with data dependency (Fig.~\ref{fig:rlworkflow} (a)). During rollout generation, rollout workers use the behavior policy, i.e., the LLM, to produce responses and collect batches of trajectories; subsequently, actor workers perform policy optimization on the accumulated fresh on-policy samples and update the behavior policy with new parameters.

Despite its effectiveness, this RL workflow introduces significant system efficiency challenges in large-scale RL training. Rollout generation is particularly expensive for LLMs, as responses are generated autoregressively and can account for around 80\% of total training time~\cite{laminar}, with multi-turn agentic RL further amplifying this cost. Moreover, response lengths and interaction turns vary substantially across prompts, resulting in severe GPU underutilization in synchronous RL training (\circled{1} in Fig.~\ref{fig:rlworkflow} (a), where trainer needs to collect a batch of on-policy data to start policy optimization). 
To mitigate these inefficiencies, recent systems increasingly adopt asynchronous RL (Fig.~\ref{fig:rlworkflow} (b)), where rollout generation and policy training proceed concurrently on disaggregated resources, and the trainer updates the policy with stale samples generated by rollout workers using earlier policy versions. By enabling trainers and rollout workers to operate in parallel, asynchronous RL execution can significantly reduce long-tail generation bubbles and improve hardware utilization. As LLM RL workloads continue to scale in terms of model size, context length, and interaction complexity, such asynchronous designs are rapidly evolving from a performance optimization into a practical requirement~\cite{glm5,minimax2026forge}.

Introducing asynchrony alone does not guarantee efficient or stable RL training, as policy staleness can degrade model performance.
Therefore, effective asynchronous RL systems must jointly balance execution efficiency, algorithmic stability, and resource utilization, raising key challenges at the intersection of distributed systems and RL algorithms. 

\vspace{1mm}
\noindent{\bf Challenge \#1: Residual data dependencies still limit system efficiency.}
Asynchronous RL allows the trainer to use trajectories generated from slightly stale policies, reducing resource under-utilization from waiting for long-tail generation. However, existing asynchronous RL systems still incur undesirable GPU idling bubbles due to residual coupling of training and generation. As shown in Fig.~\ref{fig:rlworkflow} (b), the trainer may still block until enough rollout generation data is collected to form a batch.
This phenomenon is particularly pronounced in LLM RL, where rollout time dominates overall execution time and varies substantially due to diverse response lengths and potentially complex environmental interactions. 

\vspace{1mm}
\noindent{\bf Challenge \#2: Limited system--algorithm co-optimization fails to meet the joint requirements of efficiency and stability.}
Effective asynchronous RL requires jointly optimizing system efficiency and algorithmic stability.
Stale-policy rollouts
can destabilize learning if not properly controlled, leading to
a \textit{staleness dilemma}: small staleness cannot hide long-tail generation latency~\cite{one_step_off,llamarl,asyncflow,streamrl}, while excessive staleness causes severe policy mismatch and training instability~\cite{areal,asyncrlhf}.
Although recent algorithmic correction methods can improve RL robustness, they often overlook system-level efficiency and yield only limited performance gain~\cite{gac,asyncrlhf}.

\vspace{1mm}
\noindent{\bf Challenge \#3: Non-adaptive system designs fail to accommodate dynamic and heterogeneous RL workloads.}
Compared with conventional distributed training, LLM RL workloads are more dynamic and heterogeneous.
Rollout generation is dominated by autoregressive decoding and environmental interaction, whose costs vary significantly across samples due to diverse response lengths and unpredictable tool-execution latency.
In contrast, cost of policy training is relatively regular and batch-oriented.
Moreover, these
costs evolve during training and vary across models, tasks, and hardware platforms, causing shifting resource demands of different
RL stages over time.
Non-adaptive resource allocation designs in existing asynchronous RL systems~\cite{one_step_off,verl_async,areal} are therefore inherently inefficient: fixed configurations like resource allocation and generation concurrency
can cause severe GPU load imbalance and resource under-utilization.
On the other hand, naive runtime resource reconfiguration is also expensive, as it requires cold-starting generation and training engines on new devices and rebuilding communication groups, often leaving all GPUs idle for several minutes.

\vspace{1mm}
\noindent{\bf Our Approach.}
We propose \sysname{}, a fully decoupled and adaptive asynchronous RL framework for efficient LLM post-training. \sysname{} aims to accelerate training by reducing the demand for expensive rollouts through system--algorithm co-design.
Our key insight is that, beyond widely-adopted policy asynchrony, \emph{experience replay}~\cite{lin1992self} offers an additional, underexplored opportunity to improve LLM RL efficiency.
\textbf{By carefully reusing recently generated samples, we enable experience replay in RL post-training to significantly reduce the demand for costly new rollouts without introducing excessive policy mismatch for model optimization.}
This is particularly valuable in multi-turn agentic RL, where trajectories involve long-term interactions.
We also provide theoretical guarantee that experience replay can reduce gradient variance and stabilize training.
Together, these properties enable more flexible and efficient asynchronous execution.

Based on this insight, \sysname{} is designed to fully exploit experience replay to achieve efficient LLM RL post-training:

{\em First}, \sysname{} introduces a carefully designed dataflow named \emph{dual replay data cycle} that turns experience replay into improvements in both system throughput and training convergence.
\sysname{} redesigns the dataflow and component interactions of existing asynchronous RL systems, decoupling rollout generation from policy training.
In its replay-driven cyclic dataflow, training batches are constructed from a mixture of fresh, stale, and selected replayed samples, reducing the need for costly new rollouts while keeping policy mismatch under control.
It also adjusts the rollout budget and training batch composition accordingly at each iteration to balance efficiency and stability.
To better manage the dataflow, \sysname{} inserts a replay buffer as an intermediate data layer. The buffer enables global staleness control over fresh and replayed data, and exposes extensible sampling and filtering interfaces for customized data selection, forming another replay data cycle inside experience buffer.

{\em Second}, to address dynamic resource demands of LLM RL and new workload patterns introduced by experience replay, \sysname{} employs an elastic resource pool for efficient runtime resource scaling.
By shifting the overhead of resource reallocation from system reinitialization to lightweight worker state transitions and reassignment, \sysname{} allows compute resources to be reallocated between rollout generation and policy training within seconds, ensuring efficient fresh-sample generation to reduce policy mismatch and preserve training stability.
\sysname{} further incorporates an \emph{adaptive scheduler} that continuously optimizes generation strategies and resource allocation based on replay requirements and runtime system metrics.
By adapting to diverse and evolving system demands in replay-driven RL, the scheduler sustains both high resource utilization and stable fresh-sample supply.

We deploy \sysname{} on clusters with up to 128 GPUs and evaluate it on representative single-turn reasoning and multi-turn agentic RL tasks against state-of-the-art RL systems.
Results show that \sysname{} improves training throughput by up to $2.06\times$ while maintaining comparable or even better model performance, demonstrating its scalability and efficiency.

\vspace{-4mm}
\section{Background and Motivation} \label{sec:background}

\subsection{RL for LLM Post-Training}

Reinforcement learning (RL)
for LLM post-training
adopts RL algorithms such as PPO~\cite{ppo} and its variants~\cite{grpo,remax,gspo} to optimize the policy
LLM using feedback from reward models~\cite{rlhf,rewardmodels}, rule-based evaluators~\cite{deepseekr1,grpo}, or environment interactions~\cite{webgpt,deepseek3.2}.
A typical LLM RL workflow consists of two stages: \textit{generation} and \textit{training}. During generation, the
policy model generates
responses or trajectories for a batch of prompts; in agentic settings, this stage may further involve interactions with tools, simulators, or external environments. During training, the generated trajectories are evaluated to compute learning signals, such as rewards and advantages, which are then used to update the policy model through distributed optimization~\cite{verl}.

Despite algorithmic differences, modern LLM RL systems share a common execution pattern: large-scale rollout generation using many GPU workers followed by gradient-based policy updates. This creates a heterogeneous workload:
rollout generation requires high-throughput autoregressive inference over large prompt batches,
while training relies on distributed backpropagation with intensive GPU computation and communication; reward evaluation and trajectory processing add further compute and data-movement overhead.

Recent LLM RL systems increasingly adopt asynchronous execution that allows trainers to consume trajectories produced by rollout workers using earlier policy versions. They run rollout generation and policy training on disaggregated resources in parallel, reducing long-tail generation bubbles~\cite{areal,asyncflow,asyncrlhf,laminar,streamrl,staleflow}.

\subsection{Limitations of Existing Asynchronous RL Systems} \label{subsec:limitation}

\begin{figure}[!t]
    \centering
    \begin{minipage}[t]{0.49\linewidth}
        \centering
        \includegraphics[width=\linewidth]{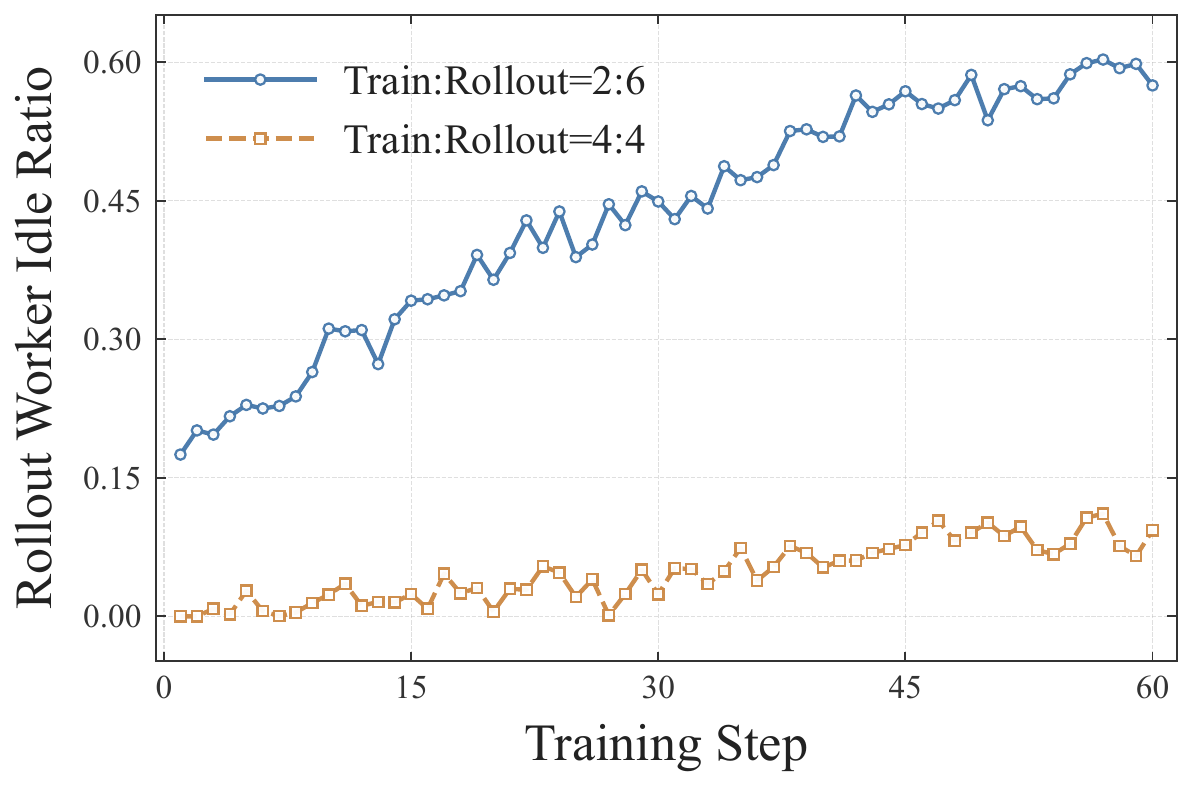}
        \caption*{(a)  Rollout idle ratio. 
        }
    \end{minipage}
    \hfill
    \begin{minipage}[t]{0.49\linewidth}
        \centering
        \includegraphics[width=\linewidth]{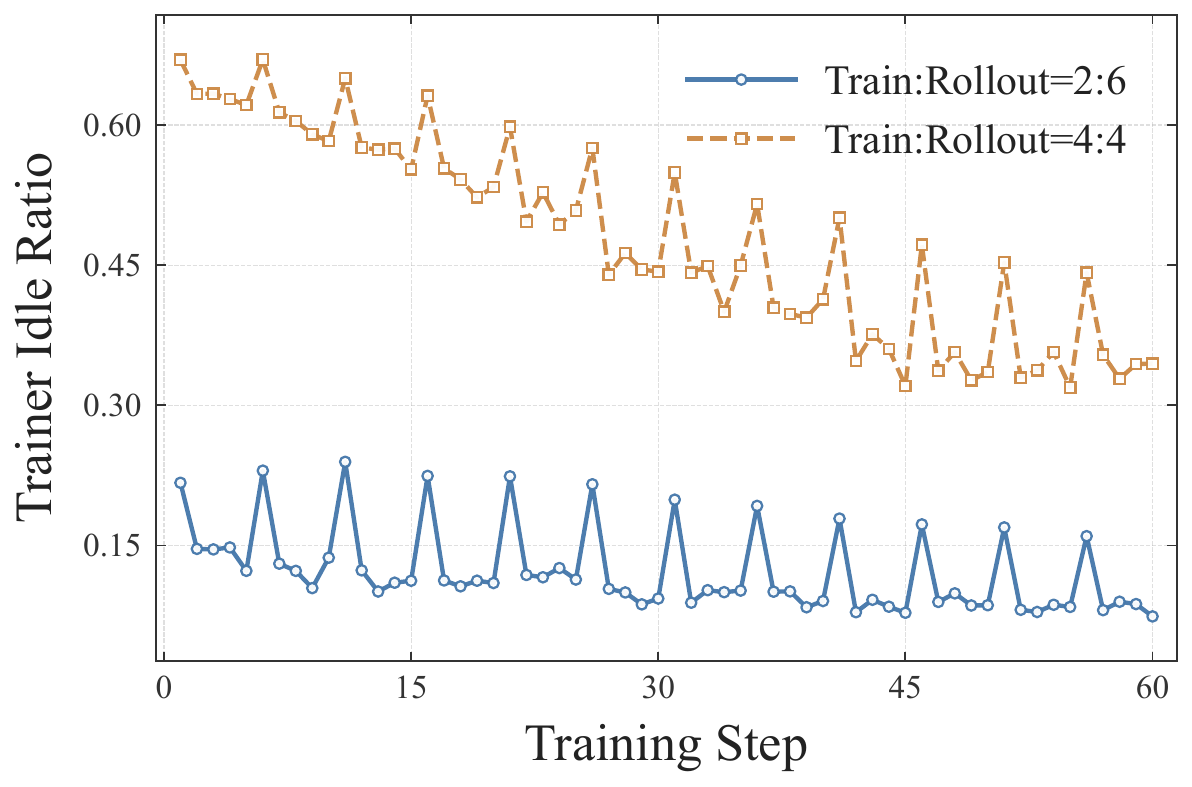}
        \caption*{(b)  Trainer idle ratio.}
    \end{minipage}

    \vspace{-2mm}
    \caption{
    GPU idle ratio (idle time / step time) under different resource allocations for Qwen3-4B training on 8 GPUs.}
    \label{fig:idle}
    \vspace{-4mm}
\end{figure}

\noindent\textbf{Strict data dependencies between training and generation.}
In
RL frameworks,
rollout generation and training usually
have strict data dependencies.

(1) \textit{Intra-stage data dependencies} within rollout generation stage,
in that
the same prompt batch is processed collectively and generated trajectories can only be consumed after the entire input batch has finished generation. Response lengths often vary substantially across prompts: generation of some trajectories terminates quickly, while others require much longer decoding. Consequently, rollout workers (GPUs) that finish early must wait for long-tail generation, resulting in significant GPU usage bubbles (\circled{1} in
Figure~\ref{fig:rlworkflow}).

(2) \textit{Inter-stage data dependencies},
that the training stage typically requires a complete batch of generated trajectories (composed from one or multiple rollout workers) for policy optimization. When rollout generation lags behind or produces insufficient samples, training GPUs are forced to stall, waiting for additional data (\circled{2} in
Figure~\ref{fig:rlworkflow}). Conversely, when the system bottleneck lies on the trainer side, rollout workers must wait for the trainers before advancing to the next generation batch.
Figure~\ref{fig:idle} shows that
the GPU idle time due to data dependencies can occupy up to 60\% of the
training iteration time,
severely limiting overall system utilization.

Existing asynchronous RL systems
address this inefficiency mainly by optimizing rollout generation to mitigate the long-tail generation effect, through load balancing across rollout workers~\cite{seer,laminar,staleflow}, generation-order scheduling~\cite{rollpacker,april}, and dynamic batching~\cite{streamrl,laminar}.
These techniques often involve additional profiling
or coordination overhead which may offset their performance gains, or
rely on assumptions that limit their generality.
StreamRL~\cite{streamrl} separates long and short samples generation into different resource groups, requiring response-length profiling and
prediction. Rollpacker and APRIL~\cite{rollpacker,april} defer long samples to form dedicated batches, which may alter the original data distribution and break iteration consistency for training~\cite{laminar}. Seer~\cite{seer} estimates request lengths from other requests in the same group,
suitable mostly for group-based RL algorithms with highly similar responses and incompatible with more general RL methods such as PPO. More broadly, systems that use response length as a proxy for rollout workload~\cite{streamrl,seer} are less effective in agentic RL, where tool execution time can dominate request latency.

Overall, the existing asynchronous RL systems optimize the rollout stage but retain the strict data dependencies between rollout workers and trainers.
This fundamentally limits their end-to-end training efficiency.

\vspace{1mm}
\noindent\textbf{Limited system--algorithm co-optimization.
}
In LLM post-training, a fundamental goal is to achieve the best model performance under limited system time/resource budget. This requires jointly optimizing two complementary dimensions: \textit{system efficiency}, in terms of how many training samples can be generated and consumed per unit time, and \textit{algorithmic convergence}, in that how effectively each sample improves the policy. We formalize the overall RL training performance of an execution design $\mathcal{S}$ as
\begin{equation}
\begin{aligned}
\mathrm{Perf}(\mathcal{S})
&=
\underbrace{\mathrm{Throughput}(\mathcal{S})}_{\text{system efficiency}}
\cdot
\underbrace{\mathrm{DataEfficiency}(\mathcal{S})}_{\text{algorithmic convergence}}
\end{aligned}
\label{eq:rl_overall_performance}
\end{equation}
$\mathrm{Throughput}(\mathcal{S})$ measures the number of valid samples generated and consumed per unit system time, while $\mathrm{DataEfficiency}(\mathcal{S})$ measures the training improvement achieved per unit data. 

\begin{figure}[!t]
    \centering

    \begin{minipage}[t]{0.48\linewidth}
        \centering
        \includegraphics[width=\linewidth]{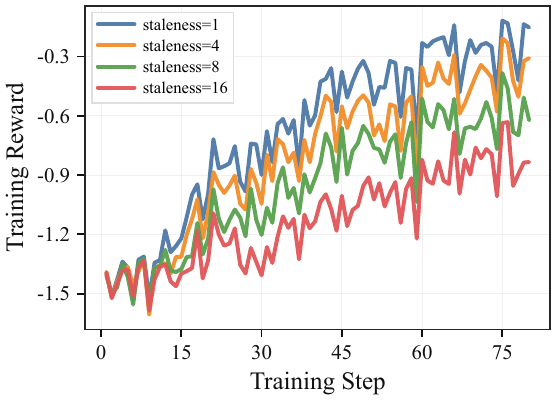}
        \caption*{(a)  Convergence.}
    \end{minipage}
    \hfill
    \begin{minipage}[t]{0.45\linewidth}
        \centering
        \includegraphics[width=\linewidth]{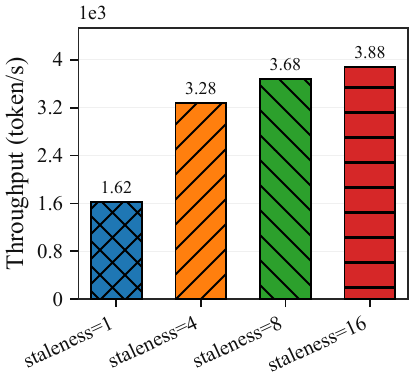}
        \caption*{(b)  Throughput.}
    \end{minipage}
    \caption{Training model performance under different staleness settings when training Qwen3-1.7B on 8 GPUs, where staleness denotes the number of training iterations between two consecutive parameter synchronizations.
    }
    \label{fig:staleness_performance}
    \vspace{-4mm}
\end{figure}

Asynchronous RL systems are facing the \textit{staleness dilemma} (Figure~\ref{fig:staleness_performance}):
increasing staleness can better decouple training and generation, thereby reducing rollout bubbles and improving their execution overlap; on the other hand, larger staleness amplifies policy mismatch, which can reduce data efficiency and destabilize training, offsetting the system-level benefits~\cite{laminar,areal}.

To balance this trade-off, many asynchronous systems adopt a conservative \textit{one-step-off} strategy~\cite{streamrl,llamarl,one_step_off,asyncflow}, that the generation policy lags behind the training policy by only one version. Some further reduce policy staleness by mixing one-step-off stale samples with on-policy samples for training~\cite{verl_async}.
While this maintains relatively fresh training data, the limited staleness window provides little room to hide long-tail rollout latency, resulting in only modest improvements in system efficiency~\cite{laminar}. Allowing larger staleness can further decouple the training pipeline and improve resource utilization, but
the system benefit of increasing staleness quickly
diminishes:
beyond a certain point, the additional overlap between generation and training becomes marginal, limiting further throughput gains.

There is another line of work focusing on improving the algorithmic robustness of asynchronous RL under high asynchrony~\cite{rolloutcorrection1,rolloutcorrection2,gac,m2po,bapo}. These approaches may stabilize training, but often overlook system efficiency implications. For instance, GAC~\cite{gac} requires additional gradient storage to handle stale updates, which significantly increases GPU memory consumption. Other works do not consider system-level optimization opportunities~\cite{asyncrlhf,m2po,bapo}. Consequently, it remains unclear whether such techniques can translate algorithmic improvements into meaningful end-to-end training performance gains in LLM RL.

Overall, neither system- nor algorithm-oriented approaches are sufficient for building an effective LLM RL framework. The key challenge is to develop a co-design that jointly accounts for system throughput and learning dynamics.

\vspace{1mm}
\noindent\textbf{Non-adaptive system designs not accommodating dynamic RL workloads well.}
Existing RL frameworks often adopt static execution strategies, in which resource allocation and key system configurations are fixed during RL training,
not adaptive to the highly dynamic workloads.

\begin{figure}[!t]
    \centering
    \includegraphics[width=\linewidth]{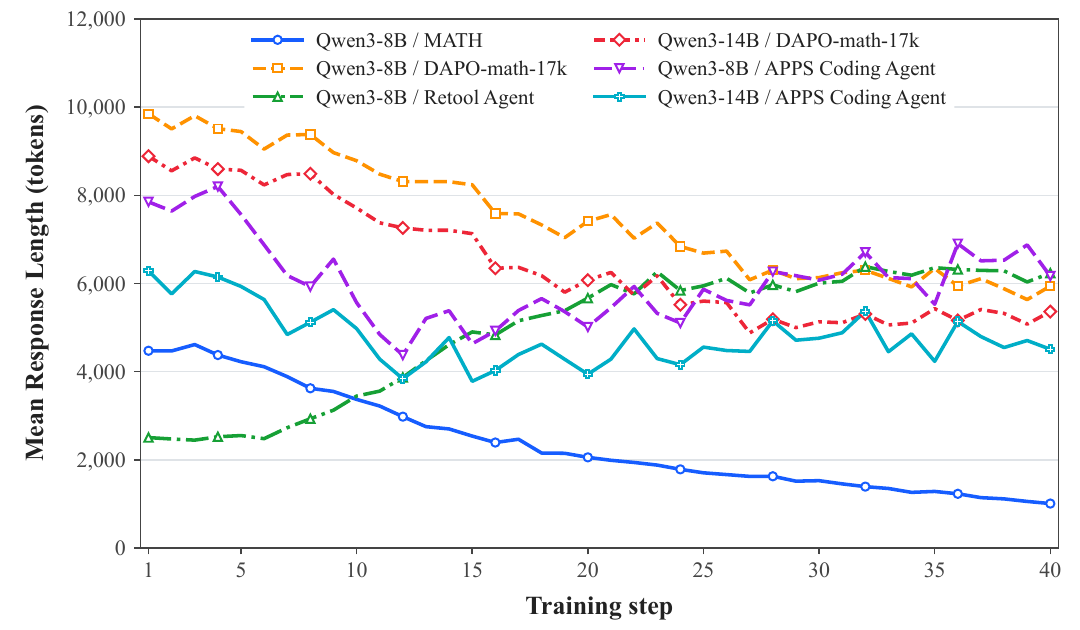}
    
    \vspace{-2mm}
    \caption{Diverse and evolving generation characteristics under different training configurations, demonstrating the highly dynamic workload pattern of LLM RL.
    }
    \label{fig:response}
    \vspace{-5mm}
\end{figure}

{\em First}, LLM RL post-training exhibits substantial \textit{runtime dynamics}. As the policy evolves, the distribution of rollout trajectories changes continuously, often resulting in shifting response length patterns (Figure~\ref{fig:response}). Early and late training steps can exhibit markedly different rollout behaviors, including variations in response length, reasoning structure, and computational intensity. Consequently, the workload composition across decoding, memory usage, and compute demand evolves throughout training, making initially selected static system configurations increasingly suboptimal over time, as shown in Figure~\ref{fig:idle}.

{\em Second}, RL workloads are highly sensitive to diverse system and algorithmic settings. The LLM post-training ecosystem is rapidly evolving across model architectures, datasets, optimization algorithms, and hardware platforms. For example, critic-free methods such as GRPO and GSPO~\cite{grpo,gspo} eliminate the need for a separate critic model, but require multiple responses per prompt, further shifting the bottleneck to rollout generation. Meanwhile, datasets vary from short-answer benchmarks such as GSM8K~\cite{gsm8k} to long-chain reasoning datasets such as DAPO-Math-17K~\cite{dapo}, leading to substantially different rollout patterns.
Emerging \textit{agentic RL} further amplifies this variability, as rollouts may involve tool calls, multi-step reasoning, and interactions with external environments, resulting in dynamic execution paths and unpredictable latency~\cite{laminar,asearch}. Consequently, optimized designs tuned to a specific stage or workload can quickly become ineffective as workload patterns shift. This makes it difficult to determine suitable configurations before training, such as GPU allocation between rollout and training or the number of concurrent generation requests.

Taken together, these factors make it difficult for a single static execution strategy to remain efficient throughout RL training. Adaptive system designs are therefore needed to dynamically adjust execution strategies as workload evolve.

\subsection{Experience Replay in Asynchronous RL}\label{subsec:experience_replay}

\begin{figure}[!t]
    \centering
    \begin{minipage}[t]{0.49\linewidth}
        \centering
        \includegraphics[width=\linewidth]{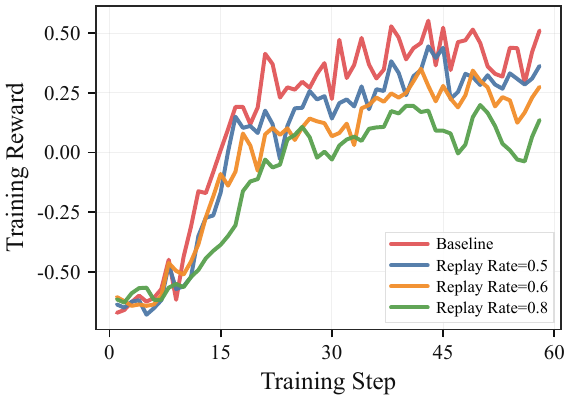}
        \caption*{(a)  Performance vs. step. 
        }
    \end{minipage}
    \hfill
    \begin{minipage}[t]{0.49\linewidth}
        \centering
        \includegraphics[width=\linewidth]{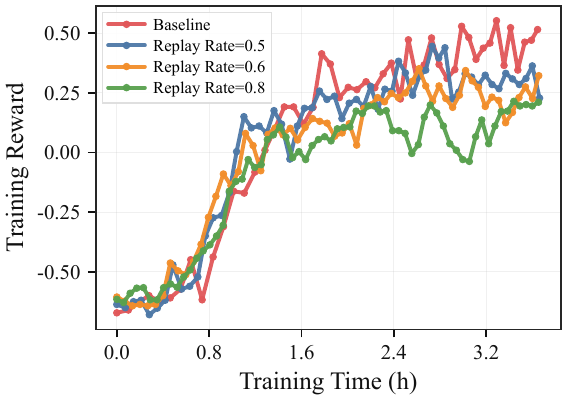}
        \caption*{(b)  Performance vs. time.}
    \end{minipage}

    \vspace{-2mm}
    \caption{
    Qwen3-4B on DAPO-MATH-17K. Naive experience replay slows convergence and degrades performance.}
 \label{fig:er_fail}
 \vspace{-4mm}
\end{figure}

Experience replay is widely used in conventional deep RL (e.g., DQN~\cite{mnih2015human}) to improve data efficiency and stabilize training~\cite{lin1992self,schaul2016prioritized,lillicrap2015continuous, fujimoto2018addressing, haarnoja2018soft}. By storing past trajectories and reusing them for multiple policy updates, it partially decouples data collection from policy optimization and amortizes the cost of expensive environmental interactions.

\vspace{1mm}
\noindent\textbf{Opportunity.}
In LLM RL post-training, where rollout generation is often dominated by long autoregressive decoding, experience replay provides a natural way to improve system efficiency. 
When newly generated trajectories are insufficient to form a batch, the trainer can continue optimization with replayed trajectories, thereby relaxing the strict dependency between rollout generation and training and mitigating long-tail rollout delays. Unlike simply increasing policy staleness in rollout generation, whose benefits quickly diminish~\cite{areal}, increasing the replay ratio directly reduces the demand for new rollouts and lowers generation workload. Moreover, we further show that introducing replay samples can help reduce gradient variance and stabilize training,
which is particularly important for LLM RL.
This makes experience replay particularly attractive for asynchronous LLM RL, with several challenges introduced.

\vspace{1mm}
\noindent\textbf{Challenge \#1: Experience replay alone does not guarantee better end-to-end training performance.}
Similar to stale samples, replayed trajectories introduce off-policy data, potentially causing distributional bias and hinder convergence. Thus, whether experience replay can improve LLM RL efficiency without compromising model convergence remains unclear.
Recent studies provide only partial evidence. RePO~\cite{repo} can outperform on-policy GRPO in some cases, but its additional updates on replayed data increase training time by around 15\%, undermining its efficiency gains. Another concurrent work~\cite{efficientrltrainingllms} analyzes how experience replay can reduce computation, but lacks system-level optimization and is evaluated at limited scale and scope.
Our experiments show that naively reusing a fixed fraction of samples from the previous step can slow convergence and degrade final model performance (Fig.~\ref{fig:er_fail}).
These highlight that replay cannot be treated as a simple source of training data. Realizing its efficiency while preserving training stability remains an open challenge.

\vspace{1mm}
\noindent\textbf{Challenge \#2: Experience replay
reshapes workload patterns and introduces additional dynamics into
LLM RL post-training.}
Increasing the proportion of replayed trajectories in each training batch reduces the rollout workload and shifts the system bottleneck toward the trainer. Combined with the evolving and highly dynamic workload characteristics described in Sec.~\ref{subsec:limitation}, it is even more difficult
to decide optimized training configurations and maintain
high efficiency throughout RL training. The optimized settings can vary significantly over time, depending on both the current system conditions and the data replay strategy.

\vspace{-2mm}
\section{\sysname{} System Overview} \label{sec:overview}

\begin{figure}[!t]
    \centering
    \includegraphics[width=\linewidth]{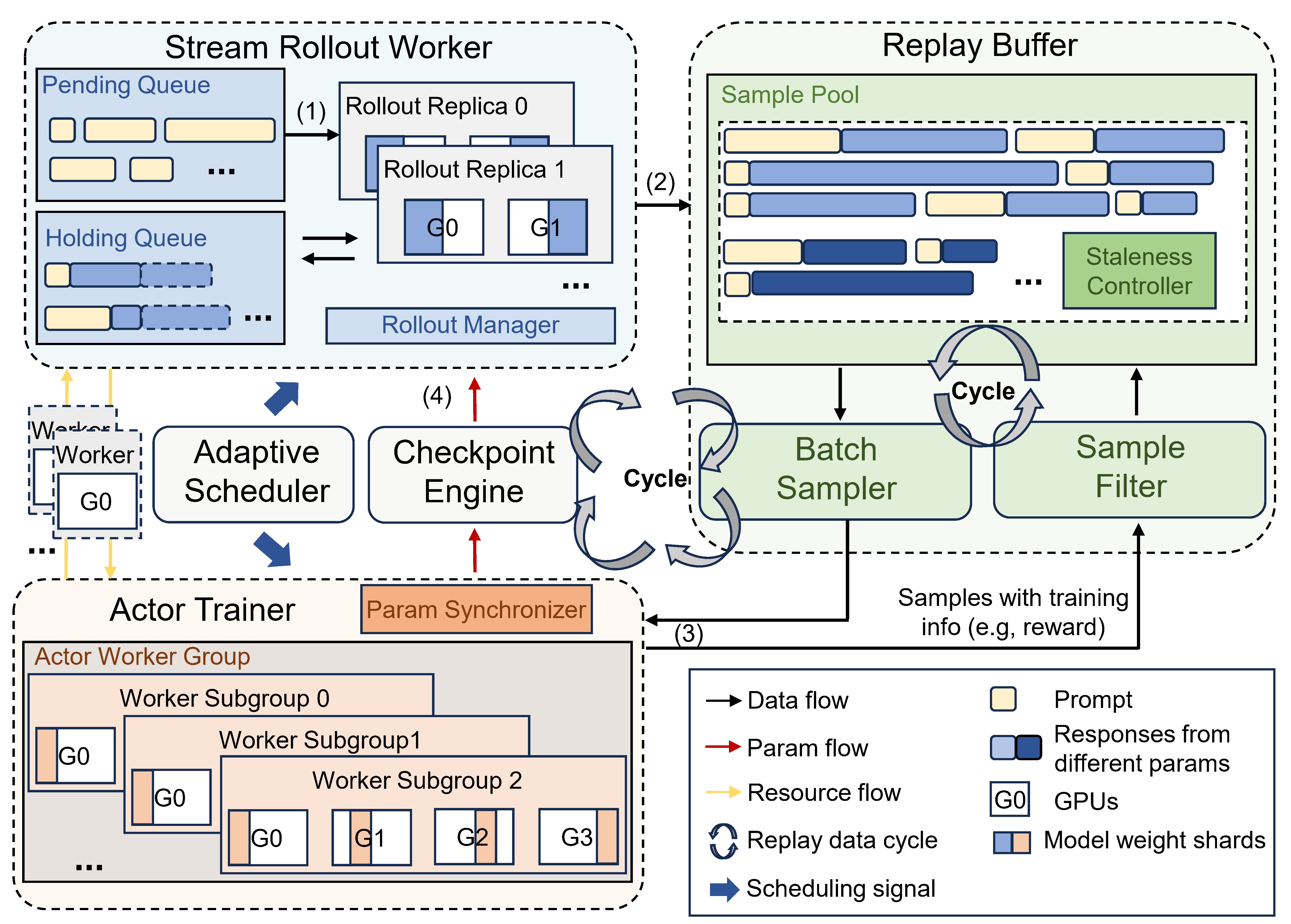}
    \vspace{-3mm}
    \caption{System overview of \sysname{}. 
    }
    \label{fig:overview}
    \vspace{-5mm}
\end{figure}

We introduce \sysname{}, an efficient LLM RL post-training framework that leverages experience replay to substantially reduce the need for costly new rollouts without incurring excessive policy mismatch.
This introduces a new optimization dimension beyond conventional model asynchrony and expands the system design space.
Building on this insight, \sysname{} adopts a fully decoupled and adaptive architecture centered on a dual replay data cycle, with system--algorithm co-design to jointly improve system efficiency and training stability.
Figure~\ref{fig:overview} presents the overall architecture of \sysname{}, which consists of four main components.

\textbf{Stream Rollout.}
The rollout worker continuously generates trajectories using the behavior policy model and interacts with the environment in agentic tasks. 
It consists of a rollout manager for request scheduling and rollout coordination, and a set of rollout workers that run serving engines such as vLLM~\cite{vllm} for inference. 
Unlike conventional batched rollout, \sysname{} adopts sample-level streaming generation, which continuously produces trajectories and pushes them to the replay buffer. We also enable support for partial rollouts~\cite{partialrollout}. This design reduces GPU idle time caused by long-tail response generation and improves rollout throughput.

\textbf{Replay Buffer.}
The replay buffer is the central data layer of \sysname{}.
It stores generated trajectories and supports replay-driven cyclic dataflow by reusing them across training iterations.
It consists of a staleness-aware sample pool, a batch sampler, and a sample filter that determines reuse eligibility.
\sysname{} exposes modular interfaces for customizing the sampler and replay filter, enabling flexible replay policies.

\textbf{Trainer.}
The trainer consumes trajectories from the replay buffer and updates the policy model using RL optimization algorithms.
It also handles other computations associated with experience consumption, such as reference model evaluation and critic model updates. The trainer can run on multiple GPUs in a distributed training setting.

\textbf{Adaptive Scheduler.}
The adaptive scheduler orchestrates the system by dynamically controlling rollout generation and resource allocation. 
Based on runtime metrics and replay requirements, it periodically adjusts generation scheduling and resource placement between rollout workers and trainers. 
This enables \sysname{} to maintain high GPU utilization while controlling policy staleness across diverse and evolving LLM RL workloads, in the presence of experience replay.

\section{Dual Replay Data Cycle}
\subsection{Replay Data Cycle in Decoupled Architecture}\label{subsec:replay_arch}
\begin{figure}[!t]
    \centering
    \includegraphics[width=0.8\linewidth]{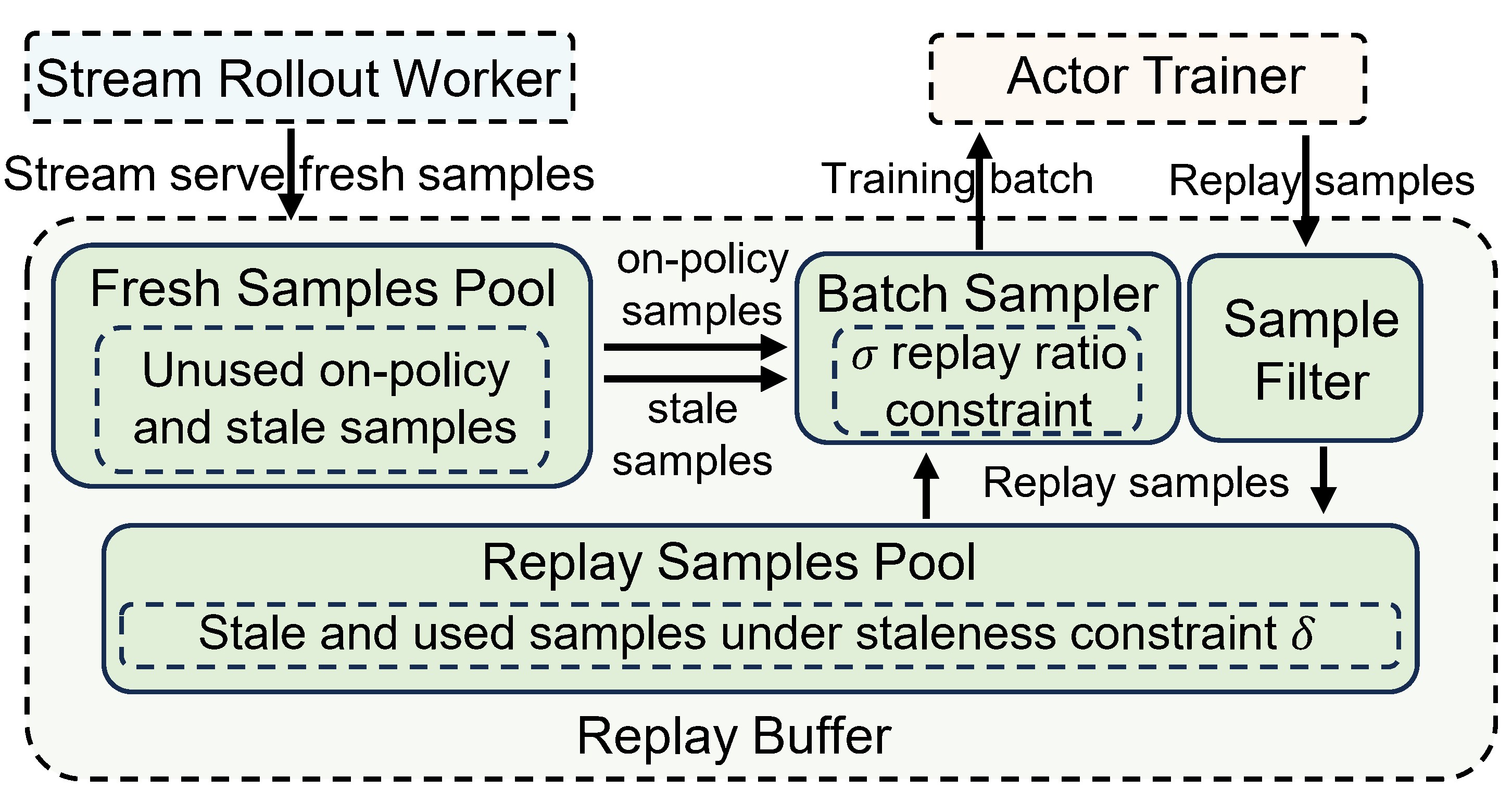}
    \vspace{-3mm}
    \caption{Dataflow of \sysname{}.
     }
    \label{fig:workflow}
    \vspace{-5mm}
\end{figure}

\sysname{} leverages experience replay to further decouple training from generation and mitigate the rollout bottleneck.
To exploit replay while addressing the stability challenges discussed in Section~\ref{subsec:experience_replay}, we design a replay-driven dataflow built on a fully decoupled architecture (Fig.~\ref{fig:workflow}).
Table~\ref{tab:replay_hyperparameters} summarizes the key hyperparameters of our replay data cycle.

\textbf{(1) Request Dispatch.}
Input prompts are first packed and dispatched to the rollout workers as sample-level requests and then inserted into a pending queue to be served in a streaming manner.
We set an upper bound $B_{\max}$ to cap the number of samples dispatched to rollout workers at each step. A larger $B_{\max}$ allows the inference engine to process more generation requests in parallel and better fill long-tail bubbles, but may also introduce more stale samples into the next policy-optimization step.

\textbf{(2) Rollout Generation.}
Rollout workers continuously generate trajectories in a sample-level streaming fashion, agnostic to training steps, batch size and model versions.
Completed trajectories are asynchronously pushed to the replay buffer.

\textbf{(3) Replay-Driven Training.}
The trainer continuously samples trajectories from the replay buffer to perform policy updates. Since the buffer contains both newly generated samples from different model versions (unused fresh on-policy and stale samples) and replay samples (trajectories used in former iteration), the trainer can construct batches from various sources, reducing its dependence on new rollouts.

As shown in Figure~\ref{fig:workflow}, rather than fixing the data replay ratio, we argue that the replay ratio should adapt to the relative rates of rollout generation and training consumption to accommodate fluctuations in sample availability.
However, unrestricted replay can exacerbate policy mismatch and hinder convergence.
We therefore introduce a \emph{data refeed boundary} $\sigma$, which caps the replay fraction while allowing it to vary with runtime conditions.
This bounded flexibility balances system efficiency and training stability, with each batch of size $b$ consuming between $b(1-\sigma)$ and $b$ unused samples.

\begin{table}[t]
\centering
\small
\caption{Hyperparameters in dual replay data cycle.}
    \vspace{-1mm}
\label{tab:replay_hyperparameters}
\begin{tabular}{@{}cp{0.78\linewidth}@{}}
\hline
Symbol & Description \\
\hline
$\sigma$ & Maximum fraction of replay data in a training batch. \\
$\tau$ & Loss weight controlling the relative contribution of replayed trajectories. \\
$\delta$ & Maximum data policy lag allowed for training. \\
\hline
\end{tabular}
    \vspace{-4mm}
\end{table}

Moreover, to mitigate potential policy shifts caused by replayed samples, we reweight its 
policy loss:
\[
w_i =
\begin{cases}
\omega_i, & i \in \mathcal{D}_{\mathrm{unused}},\\
\tau \omega_i, & i \in \mathcal{D}_{\mathrm{replay}},
\end{cases}
\qquad
\tilde{w}_i = \frac{w_i}{Z},
\quad
Z = \frac{\sum_j w_j n_j}{\sum_j n_j}.
\]
Here, $\mathcal{D}_{\mathrm{unused}}$ and $\mathcal{D}_{\mathrm{replay}}$ denote sets of unused and replay data, $\tau \ge 0$ is the replay loss weight, $\omega_i$ is the importance weight, and $n_i$ is the number of valid tokens. The normalized weight $\tilde{w}_i$ preserves the policy-loss scale while keeping replayed samples relatively down-weighted.

\textbf{(4) Parameter Synchronization.}
To control policy staleness, the trainer triggers parameter synchronization with rollout workers through a checkpoint engine after policy updates every few training steps.
Due to the stream dispatch design in Step~(1), rollout workers may still hold several partially generated samples when receiving model synchronization signals. These requests are moved to a holding queue and continued after parameter synchronization, forming partial rollout~\cite{partialrollout}.

Overall, our replay-driven workflow combines bounded sample reuse, normalized loss reweighting, and explicit staleness control. It reduces reliance on new rollouts without simply enlarging the staleness window, enabling system--algorithm co-optimization.

\textbf{Theoretical Justification.}
\sysname{} combines two complementary mechanisms: replay reduces rollout demand,
while normalized reweighting controls replay's contribution to policy
updates. The following local analysis provides a theoretical basis for
this design. Consider a fixed replay fraction $\lambda_t\in(0,\sigma]$,
with $\sigma<1$, and equal importance-weighted token mass per sample.
Let $\hat g_t^F$ and $\hat g_t^R$ denote gradient estimates from unused
and replayed data. For $0\le\tau\le1$, normalized reweighting gives
\begin{align}
\hat g_t^{\mathrm{rw}}
&=(1-\alpha_t)\hat g_t^F+\alpha_t\hat g_t^R,
\nonumber\\
\alpha_t
&=\frac{\tau\lambda_t}{1-\lambda_t+\tau\lambda_t}
\le \frac{\tau\sigma}{1-\sigma+\tau\sigma}.
\label{eq:main_effective_replay_weight}
\end{align}
Thus, \sysname{} separately controls \emph{how much data is reused} through
$\lambda_t$ and \emph{how strongly replay influences the update} through
$\alpha_t$, allowing efficient reuse with moderated gradient influence.

\emph{Controlling gradient error.}
Let $g_t^\star=\nabla\mathcal J_t(\theta_t)$ be the local on-policy
gradient. In this simplified model, suppose $\hat g_t^F$ is unbiased
for $g_t^\star$, while $\hat g_t^R$ has bias norm at most $\Delta_t$.
Assume their centered errors are uncorrelated, with variances $v_t^F$
and $v_t^R$. All moments condition on the learner state, batch
composition, and replay weight. Then
\begin{align}
\mathbb E_t\|\hat g_t^{\mathrm{rw}}-g_t^\star\|^2
&\le \mathcal E_t(\alpha_t),
\label{eq:main_replay_error}\\
\mathcal E_t(a)
&:=\underbrace{(1-a)^2v_t^F+a^2v_t^R}_{\text{variance}}
+\underbrace{a^2\Delta_t^2}_{\text{squared bias bound}}.
\nonumber
\end{align}
For an $L$-smooth surrogate, the update
$\theta_{t+1}=\theta_t+\eta_t\hat g_t^{\mathrm{rw}}$ with
$0<\eta_t\le1/L$ satisfies
\begin{align}
&\mathbb E_t[
\mathcal J_t(\theta_{t+1})-\mathcal J_t(\theta_t)]
\nonumber\\
&\qquad\ge
\frac{\eta_t}{2}
\left(\|g_t^\star\|^2-\mathcal E_t(\alpha_t)\right).
\label{eq:main_replay_progress}
\end{align}
This directly connects replay reweighting to optimization quality:
a smaller gradient-error bound yields a stronger certified one-step
progress bound.

\emph{Benefit of weighted replay.}
For fixed source estimates and $v_t^F>0$, the error bound is minimized
over $a\in[0,\lambda_t]$ by
\begin{equation}
\alpha_t^{\mathrm{opt}}
=\min\left\{
\lambda_t,\,
\frac{v_t^F}{v_t^F+v_t^R+\Delta_t^2}
\right\}.
\label{eq:main_optimal_replay_weight}
\end{equation}
Importantly, $\alpha_t^{\mathrm{opt}}>0$ and
$\mathcal E_t(\alpha_t^{\mathrm{opt}})<\mathcal E_t(0)=v_t^F$.
Under the stated assumptions, appropriately weighted replay therefore
achieves a strictly tighter gradient-error bound than discarding replay,
without requiring additional unused samples.
Intuitively, a small replay contribution reduces the unused-data variance
term to first order, while replay variance and squared bias enter at
second order. When $\alpha_t^{\mathrm{opt}}<\lambda_t$, down-weighting
also strictly improves the bound over uniform weighting. These results
support retaining replay while adjusting its influence to balance
variance reduction and policy mismatch.

\emph{Support for our co-design.}
\sysname{}'s controls address complementary aspects of this balance.
The replay cap $\sigma$ guarantees at least $b(1-\sigma)$ unused samples
per batch, while $\tau$ moderates replay influence without changing
batch composition. For example, $\sigma=\tau=0.5$ permits half the
samples to be replayed while limiting $\alpha_t$ to $1/3$ in the
equal-mass model. Synchronization and the policy-lag limit restrict
trajectory age, helping control $\Delta_t$ when per-update policy
drift is bounded.

\vspace{-2mm}
\subsection{Replay Data Cycle in Experience Buffer}\label{subsec:replay_buffer}

Experience replay introduces a substantially more complex dataflow and data-scheduling space than existing RL systems.
\sysname{}'s experience buffer represents a central data layer enabling explicit lifecycle management for training data under experience replay.
Especially, the experience buffer is designed with two key considerations: (1) it should allow flexible data reuse, controllable staleness, and adaptive sample selection, to fully exploit replay opportunities; (2) it should be easily customizable, exposing sufficient interfaces to support algorithm-specific sampling and filtering strategies. To this end, the buffer implements a replay-driven cyclic dataflow with structured data management.

\vspace{1mm}
\noindent\textbf{Sample Pool with Staleness Control.}
The sample pool maintains fresh samples from on-policy and stale model versions, and replayed trajectories available for training.
To prevent excessive policy mismatch caused by stale data, \sysname{} enforces a global staleness $\delta$ on stored trajectories, i.e., only samples generated in the recent $\delta$ steps are available.
We enable this by
associating each sample
with additional metadata including its policy version for generation and user-customized training information (e.g, advantages).

\vspace{1mm}
\noindent\textbf{Batch Sampling.}
The sampler selects trajectories from the sample pool to construct training batches, with replay proportion in each batch determined by runtime conditions and sampling strategies and is capped by $\sigma$ (Sec.~\ref{subsec:replay_arch}). 
Users can implement customized sampling strategies to further improve training stability. For example, a medium-reward sampler selects only trajectories with intermediate rewards, i.e., trajectories that are neither too easy nor too hard~\cite{dapo}.

\vspace{1mm}
\noindent\textbf{Replay Filtering and Reinsertion.}
After being consumed by training, trajectories are passed through a replay filter that determines whether they should be reinserted into the sample pool for future reuse.
The filtering policy can incorporate quality signals such as rewards, advantages, or other algorithm-dependent metrics.
Similar to sampling, replay filtering is orthogonal to the system design and can be customized using different algorithms and replay strategies. 
To ensure a fair comparison with other RL systems, we use random sampling and accept all replay samples without filtering rather than carefully designed strategies in our experiments.

Together, the sample pool, sampler, and replay filter form a closed-loop replay data cycle.
Newly generated trajectories enter the sample pool, are sampled for training, and may be reinserted for later reuse.

\section{Dynamic Resource Placement}\label{sec:dynamic_resource}

\begin{figure}[!t]
\centering

\begin{minipage}[t]{0.43\columnwidth}
\vspace{3mm}
\centering
    \small
    \begin{tabular}{lc}
        \toprule
        Settings & Cost \\
        \midrule
        8B / 32 GPUs   & 228.20s \\
        14B / 64 GPUs  & 286.31s \\
        32B / 128 GPUs & 368.64s \\
        \bottomrule
    \end{tabular}
    \vspace{1mm}

    {\small (a) Total overhead.}
\end{minipage}
\hfill
\begin{minipage}[t]{0.5\columnwidth}
    \vspace{0pt}
    \centering
    \includegraphics[width=\linewidth]{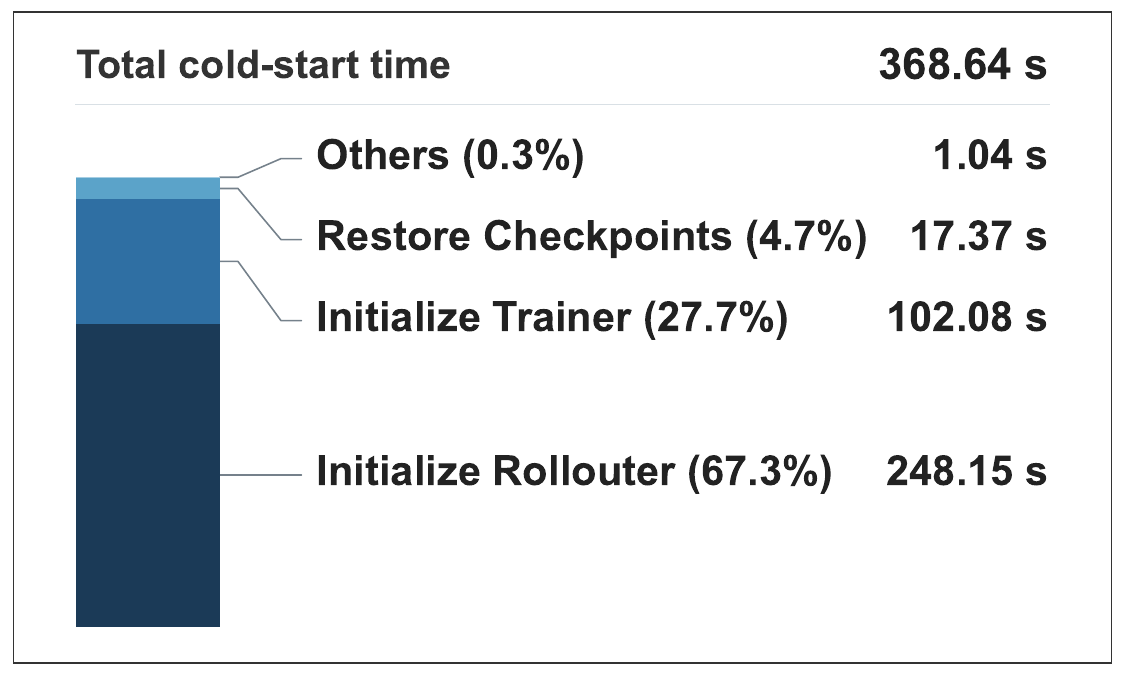}

    {\small (b) Overhead breakdown.}
\end{minipage}

\vspace{-1mm}
\caption{Cold-start overhead.
(a) Total overhead under Qwen3 models.
(b) Breakdown for Qwen3-32B on 128 GPUs.} 
\label{fig:cold_start_overhead}
\vspace{-4mm}
\end{figure}

\subsection{Design Consideration}
Dynamic RL workloads and experience replay jointly reshape the
resource demands of rollout generation and policy training.
To accommodate these evolving bottlenecks, \sysname{} requires
an elastic resource pool that efficiently reallocates GPUs
between the two stages.
Runtime scaling is expensive in asynchronous RL systems due to substantial system overhead.
Rollout workers host inference engines with large KV caches, whereas training workers maintain optimizer states, gradients, and training model shards. Training and rollout often rely on different parallelism strategies, operator implementations, and fixed communication groups. 
Resource scaling under disaggregated placement is fundamentally more complex than the fixed training-generation context switch in synchronous RL systems such as VeRL~\cite{verl}. Rather than switching context under a stable topology, the distributed execution structure itself needs to be changed, requiring communication-group reconfiguration, resource remapping, and materialization of missing
states such as up-to-date optimizer states.
Reallocating GPUs across stages requires expensive reinitialization and incurs significant cold-start overhead, including rebuilding inference engines, reconstructing training meshes, and relaunching distributed jobs—operations that can consume minutes of GPU time, as shown in Figure~\ref{fig:cold_start_overhead}.

\subsection{Elastic Resource Pool Using Unified AsyncHybrid Worker}
To eliminate heavy cold-start overhead in runtime resource placement, \sysname{} introduces the \emph{Unified AsyncHybrid Worker}, a unified execution abstraction that decouples worker lifetime from its assigned role.
Instead of maintaining separate rollout and training workers like existing systems~\cite{laminar,one_step_off,verl_async,areal}, our Unified AsyncHybrid Worker encapsulates necessary runtime modules for both inference and training, and can dynamically switch roles on demand, enabling flexible
resource reallocation.
To mitigate the additional memory overhead for maintaining dual capabilities, \sysname{} employs a selective offloading strategy, by which inactive modules (e.g., inference model when training) are temporarily offloaded to CPU memory, thereby preserving valuable GPU memory for the active workload. 

\sysname{} treats workers (each holding a GPU)
as the basic unit of resource management.
Resource pools of high-level abstractions, such as the Stream Rollout Component and the Trainer, are defined as collections of Unified AsyncHybrid Workers, and runtime scaling is achieved by adjusting GPU assignments to run the workers and switching execution states.
By replacing heavyweight system reinitialization with lightweight worker reassignment and state transitions, \sysname{} enables efficient and flexible elastic scaling under evolving workload demands.

\begin{figure}[!t]
    \centering
    \includegraphics[width=\linewidth]{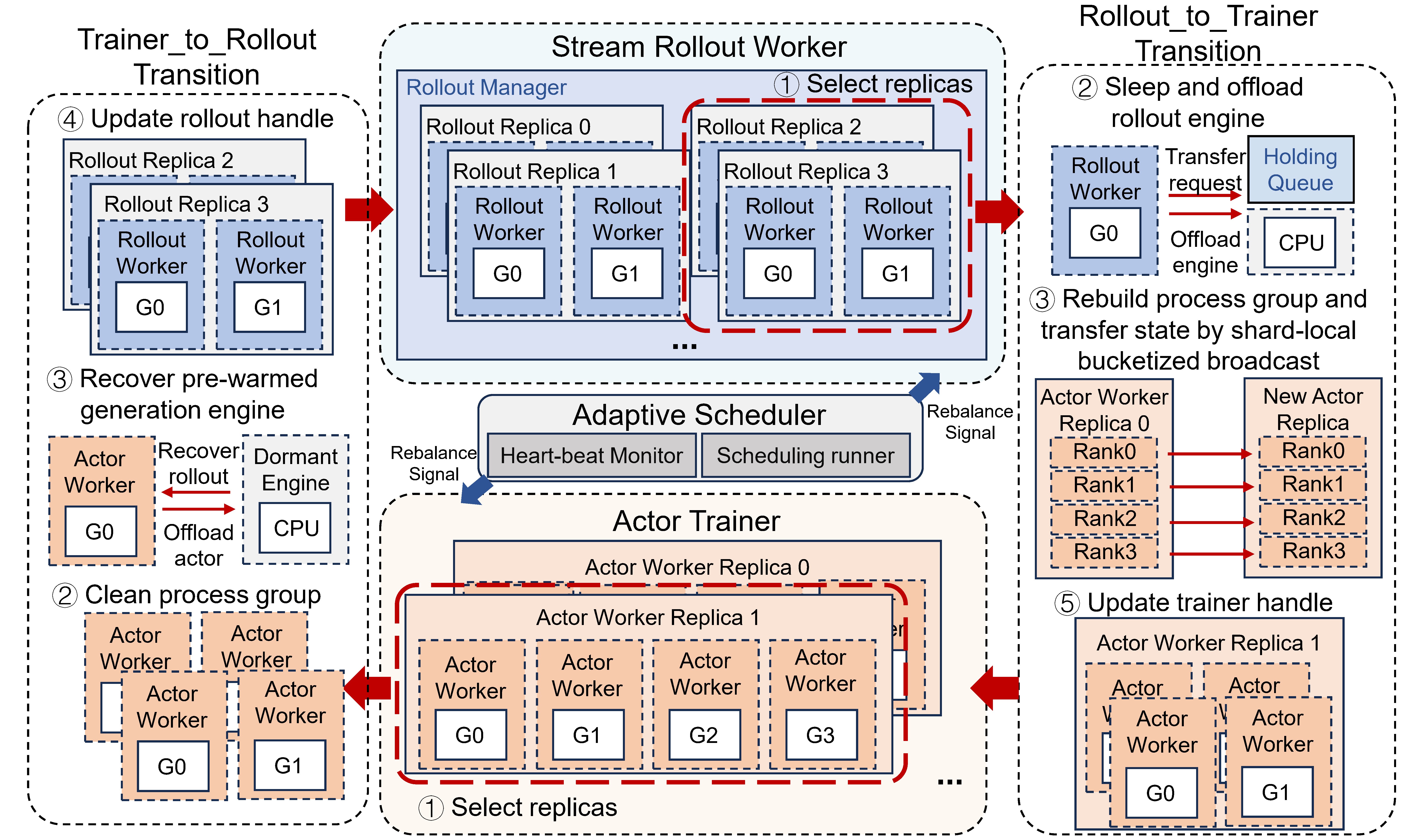}
    \caption{Workflow of \sysname{}'s
    resource rebalance. We use TP=2 for the rollout worker and FSDP group size=4 for the trainer as an example.
    }
    \label{fig:drp_process}
    \vspace{-4mm}
\end{figure}

\vspace{-3mm}
\subsection{Resource Rebalancing}\sysname{} builds an elastic resource pool on top of the Unified AsyncHybrid Worker abstraction to support efficient resource rebalancing between rollout generation and policy training. The rebalancing workflow is given in Figure~\ref{fig:drp_process}, whose design follows two principles. {\em First}, resources are reassigned in complete replica groups
to preserve valid parallelism configurations.
Let $s_T$ and $s_R$ denote the GPUs per trainer and rollout replica. For a pool of $W$ GPUs,
\begin{equation}
A=n_Ts_T,\qquad R=n_Rs_R,\qquad A+R=W,
\label{eq}
\end{equation}
where $n_T$ and $n_R$ are the corresponding replica counts. {\em Second}, shard partitions are preserved when possible, allowing retained trainers to keep local model and optimizer shards and avoiding costly full-state gathering or repartitioning.

\vspace{1mm}
\noindent\textbf{Pre-Warm Initialization.}
To avoid cold-start reinitialization overhead, we perform a pre-warm stage before RL training for expensive runtime contexts such as inference engines.  
At startup of training, workers are initialized as hybrid workers capable of serving both trainer and rollout roles. Each worker may hold both the training and inference engines, while inactive modules are offloaded or put into a dormant state to reduce GPU memory usage. This amortizes expensive setup costs during resource rebalance such as process initialization, CUDA graph capture, and inference engine construction.

\vspace{1mm}
\noindent\textbf{Role Transition.}
Lightweight bidirectional role transitions between trainer and rollout pools are enabled, by preserving shard layouts and reusing pre-warmed components. When some resources are shifted from trainer workers to rollout workers, shifting workers activate their dormant rollout modules, while the remaining trainer workers keep their local parameter and optimizer shards.
Since only the replica dimension changes, no full model or optimizer state needs to be gathered or redistributed. Newly added rollout workers then receive up-to-date policy model weights.
When resources are shifted
from rollout workers to trainer workers, \sysname{} similarly avoids full-state materialization by initializing new trainer replicas through shard-local broadcast: existing trainer replicas serve as sources, and each worker transfers only its local parameter and optimizer shard to the corresponding worker in the new replica. We implement bucketized CUDA broadcasts optimization, which is inspired by Pytorch DDP~\cite{pytorchddp} to reduce collective launch overhead. These shard-local transfers proceed in parallel and avoid full gather, CPU staging, or global redistribution. Together, pre-warmed rollout reuse and shard-local trainer initialization
enable efficient resource rebalancing as local state changes rather than costly subsystem restart.

Moreover, we typically 
trigger resource rebalancing upon periodical parameter synchronization (Sec.~\ref{subsec:replay_arch}), avoiding conflicts between inference and policy update
and extra parameter updates for new
rollout workers. After the transition completes, \sysname{} updates the trainer and rollout worker-group handles and resumes normal parameter synchronization.

\section{Adaptive Scheduling for Dynamic RL Workloads}
\label{sec:adaptive_scheduler}

We further design an adaptive scheduler for short-term generation scheduling among rollout workers and long-term resource rebalancing across trainer and rollout workers, adapting to both generation dynamics and resource-demand shifts. \vspace{-2mm}
\subsection{Adaptive Generation Scheduling}

Our sample-level streaming rollout largely removes the long-tail generation bottleneck, shifting the main challenge to \emph{generation concurrency control} (which decides how many requests a rollout worker receives at a time) and {\em load balancing} among rollout workers.  
Low concurrency under-utilizes rollout GPUs, while aggressive concurrency for long-context workloads can create KV-cache pressure and excessive preemption~\cite{seer}. Our adaptive generation scheduling keeps rollout workers highly utilized while avoiding KV-cache congestion, exploiting: a slow feedback controller that adjusts the real-time effective concurrency cap of each rollout worker every few seconds, and a fast global request dispatcher that admits and routes incoming requests to rollout workers subject to these caps.

\vspace{1mm}
\noindent\textbf{Adaptive Cap Control.}
The controller periodically collects KV-cache utilization, waiting queue length, generation throughput, number of preemption requests and active generation tasks of each inference server. For agentic workloads, only active LLM generations are counted toward concurrency. Each worker $i$ maintains EWMA-smoothed KV utilization ($\hat{u}_{i,t}$) and preemption signals ($\widehat{\Delta P}_{i,t}$):
\begin{equation}
\hat{u}_{i,t} = \alpha u_{i,t} + (1-\alpha)\hat{u}_{i,t-1}, \quad
\widehat{\Delta P}_{i,t} = \alpha \Delta P_{i,t} + (1-\alpha)\widehat{\Delta P}_{i,t-1}. \notag
\end{equation}
The effective sequence cap $C_{i,t}$ is updated using AIMD: it increases additively when KV utilization is below the target kv cache usage band and demand exists, and decreases multiplicatively when congestion appears, such as sustained KV preemption growth or KV utilization reaching a hard limit. 

\vspace{1mm}
\noindent\textbf{Token-aware Request Admission and Routing.}
Within the adaptive caps, the global dispatcher admits requests based on projected KV-token pressure rather than request count. A request $r$ can be assigned to worker $i$ only if
\begin{equation}
\frac{T^{\mathrm{pred}}_{i,t}+\widehat{T}(r)}{Budget_i} \leq \rho, \notag
\end{equation}
where $T^{\mathrm{pred}}_{i,t}$ is the predicted active token load on worker $i$, using the average response length of recent requests to estimate workload, $\widehat{T}(r)$ is the estimated prompt-plus-response footprint and $Budget_i$ is the worker token budget. Requests are not routed to a rollout worker if observed KV utilization exceeds the admission limit of the worker, even when available request slots remain.
Among feasible rollout workers, the dispatcher chooses the one with the minimum cost:
\begin{equation}
\mathrm{cost}_i(r)
=
\max\left(
\frac{T^{\mathrm{pred}}_{i,t}+\widehat{T}(r)}{Budget_i},
\hat{u}_{i,t}
\right)
+
\lambda_p \widehat{\Delta P}_{i,t}. \notag
\end{equation}
Here $\lambda_p$ is a hyper-parameter. This favors workers with lower KV load, shorter queues, and lower preemption pressure. We further explore prompt affinity, by placing sibling requests of the same prompt on the same worker when possible to exploit prefix cache, while allowing spillover under overload.

Overall, this scheduling raises concurrency when utilization is low, prevents KV overcommitment through token-aware admission, and balances load across rollout workers under diverse, dynamic workloads.

\subsection{Rollout/Trainer Resource Allocation}
\label{sec:long_term_resource_scheduling}

Dynamic resource allocation between trainer and rollout workers is necessary to tackle time-varying workloads in asynchronous RL training. Experience replay further introduces a tradeoff between generating fresh samples and reusing buffered data. We design a lightweight resource allocation scheme adapting rollout/train resources according to the observed producer--consumer dynamics.

\begin{algorithm}[t]
\small
\caption{Rollout/Train Resource Scheduler}
\label{alg:long_term_resource_scheduler}

\KwIn{Current resource placement $(A,R)$ of Actor Trainer $A$ and Stream Rollout Worker $R$, feasible placement set $\mathcal{P}$, total workers $W$, Runtime statistics: number of produced and consumed samples $(\Delta N_{\mathrm{prod}}, \Delta N_{\mathrm{cons}})$, trainer and rollout worker idle ratio $i_{\mathrm{rollout}}, i_{\mathrm{train}}$, time window $\Delta t$ }
\KwIn{Target refeed ratio $\sigma$, small constant $\epsilon$}
\KwOut{Updated rollout and trainer worker placement}

Estimate active-time rates using idle ratio correction:
\[
\lambda_{\mathrm{r}} \leftarrow
\frac{\Delta N_{\mathrm{prod}}/\Delta t}
     {1-i_{\mathrm{rollout}}+\epsilon},
\qquad
\lambda_{\mathrm{d}} \leftarrow
\frac{\Delta N_{\mathrm{cons}}/\Delta t}
     {1-i_{\mathrm{train}}+\epsilon}.
\]

Set $q^\star \leftarrow 1-\sigma$\;

Select the best feasible placement:
\[
(A^\star,R^\star)
\leftarrow
\arg\min_{\substack{(A',R')\in\mathcal{P}\\ A'+R'=W}}
\left|
\log
\frac{
\frac{R'}{R}\lambda_{\mathrm{r}}
}{
\frac{A'}{A}\lambda_{\mathrm{d}}(q^\star+\epsilon)+\epsilon
}
\right|.
\]

\If{the cooldown gate is active}{
    \Return{$(A,R)$}\;
}
\Return{$(A^\star,R^\star)$}\;

\end{algorithm}

\noindent\textbf{Producer-consumer Modeling.} As detailed in Algorithm~\ref{alg:long_term_resource_scheduler}, we consider rollout workers as sample producers and trainers as sample consumers. The scheduler measures production, consumption, and GPU idle ratios over a recent time window (i.e., a training step) and estimates the effective rollout capacity and trainer demand. This correction prevents the scheduler from misinterpreting blocked execution caused by sample starvation or staleness constraints as intrinsic throughput changes.

\vspace{1mm}
\noindent\textbf{Scheduling Strategy.} As given in Algorithm~\ref{alg:long_term_resource_scheduler}, the scheduler searches over a discrete feasible placement set that respects global worker limits and system constraints such as tensor parallelism and trainer shard grouping. Each candidate placement is evaluated by extrapolating production and consumption rates from the current allocation. The goal is to match the predicted fresh-data coverage of the new placement plan to the target coverage induced by the desired refeed ratio $\sigma$. We use a logarithmic mismatch metric to make the decision invariant to multiplicative scale and symmetrically penalizes under- and over-supply of fresh samples.

To avoid frequent resource reallocation, a new placement is applied only when permitted by a cooldown gate, which enforces a minimum number of training steps between two consecutive rebalancing operations. 

\section{Evaluation}

\subsection{Implementation}\label{sec:implementation}
We implement \sysname{} on top of VeRL~\cite{verl} with a total codebase of 10.6K lines of Python.

\vspace{1mm}
\noindent\textbf{Fully decoupled RL architecture.} Following VeRL, we use \textsc{Ray}~\cite{ray} as distributed framework. The core implementation (6.6K LoC) realizes a fully decoupled RL architecture with a dual replay data cycle, separating rollout generation and policy training into independent pipelines. The stream rollout worker is built on \textsc{vLLM}~\cite{vllm} and the trainer is implemented with \textsc{PyTorch} using FSDP~\cite{fsdp}, while the overall design remains compatible with alternative inference and training engines. We use \textsc{NCCL} with a customized \emph{CheckpointEngine}~\cite{checkpointengine} for efficient parameter synchronization.

\vspace{1mm}
\noindent\textbf{Adaptive scheduler}
is implemented with 1.3K LoC and constructed when initializing trainer and rollout workers.
\subsection{Experiment Setup}\label{sec:evaluation}

\textbf{Testbed.}
We deploy \sysname{} on an H800 cluster with 16 nodes (128 GPUs in total). Each node is equipped with eight H800 GPUs connected by NVLink, and nodes are interconnected with 8$\times$400~Gbps bandwidth. The key software versions include CUDA~12.6, PyTorch~2.7.1, and NCCL~2.26.2.

\begin{figure*}[!t]
    \centering

    \begin{minipage}{0.48\linewidth}
        \centering
        \includegraphics[width=\linewidth]{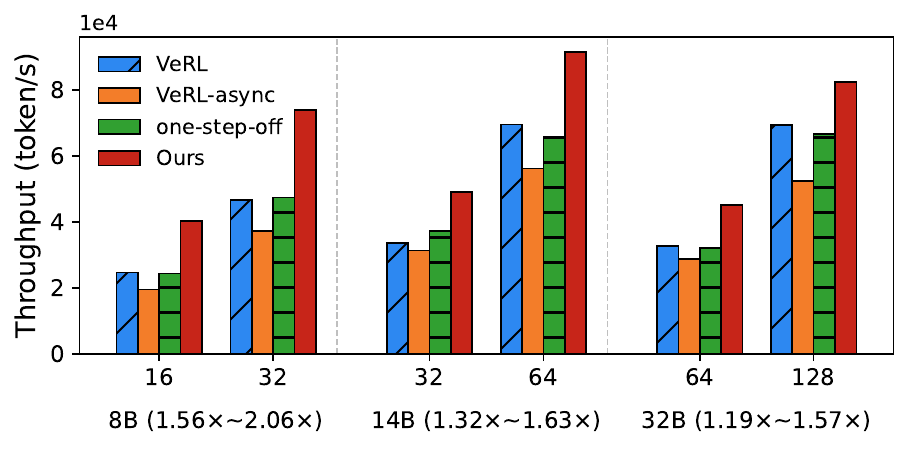}
        \centerline{(a) Single-turn reasoning RL on different number of GPUs.}
    \end{minipage}
    \hfill
    \begin{minipage}{0.48\linewidth}
        \centering
        \includegraphics[width=\linewidth]{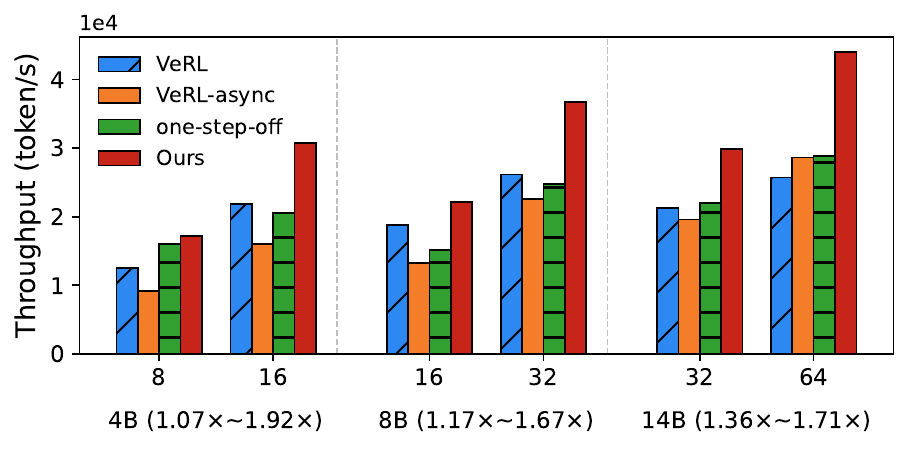}
        \centerline{(b) Multi-turn coding agentic RL on different number of GPUs.}
    \end{minipage}

    \caption{System throughput under different RL workloads.}
    \label{fig:throughput}
    \vspace{-4mm}
\end{figure*}

\vspace{1mm}
\noindent\textbf{Models and Datasets.}
We run RL training on Qwen3-4B, 8B, 14B and 32B~\cite{qwen3}, which are widely used in both research and industry. 
For single-turn reasoning RL, we follow~\cite{laminar,dapo} to use DAPO-Math-17K~\cite{dapo}, a representative benchmark for mathematical reasoning. 
For multi-turn agentic workloads, we follow DiDPO~\cite{didpo} and train coding agents on APPS~\cite{apps}, where the agent iteratively develops a solution program for challenging competitive-programming problems.
We further follow ReTool~\cite{retool} to evaluate tool-interactive reasoning on DAPO-Math-17K, where the model invokes a code sandbox during problem solving.

\vspace{1mm}
\noindent\textbf{Baselines.}
We compare \sysname{} with several state-of-the-art open-source LLM RL systems. For the synchronous RL baseline, we use VeRL~\cite{verl}, a widely adopted LLM post-training framework, with its official implementation. For asynchronous RL baselines, many concurrent systems
do not release code~\cite{laminar,staleflow}, provide limited hardware support~\cite{asyncflow}, or lack mature system optimizations~\cite{asyncrlhf}. Therefore, following~\cite{staleflow,laminar}, we use two state-of-the-art open-source asynchronous RL systems, one-step-off~\cite{one_step_off} and VeRL-async~\cite{verl_async}, as baselines. We follow~\cite{laminar} and official running scripts of baselines to set the hyper-parameters (i.e., parallelism).
To ensure fair comparison, all systems use the same software environment, with vLLM as the generation backend and FSDP2 as the training backend. 

\vspace{1mm}
\noindent\textbf{Settings.}
For design parameters in \sysname{}, we set the maximal generated samples $B_{max}$ equal to the training batch size $b$. For training refeed boundary $\sigma$ and replay loss weight $\tau$, we set both to 0.5 to reduce the policy mismatch introduced by experience replay. The global data staleness $\delta$ is set to 1, allowing only samples generated using the latest and one-version-early models to be kept in the replay buffer.

\vspace{1mm}
\noindent\textbf{Metrics.}
We measure system efficiency using throughput (tokens per second), computed as the number of tokens processed by the trainer in one step divided by the step time. We report the average throughput over five measured steps after five warm-up steps. For coding agent experiments, where complex interactions lead to higher performance variance, we average throughput over ten measured steps after ten warm-up steps.
We also evaluate average training reward during the training steps, which serves as an indicator of training stability and algorithmic effectiveness of the RL framework.

\vspace{-4mm}
\subsection{Training Throughput}

We evaluate the training throughput of \sysname{} against various baselines across single-turn reasoning and multi-turn agentic workloads, varying both model size and cluster scale.

\noindent\textbf{Single-Turn Reasoning.}
As shown in Figure~\ref{fig:throughput}~(a), \sysname{} consistently achieves the highest throughput for Qwen3-8B, 14B, and 32B on 16--128 GPUs, delivering $1.19\times$--$2.06\times$ speedups over the evaluated baselines.
The speedups are $1.56\times$--$2.06\times$, $1.32\times$--$1.63\times$, and $1.19\times$--$1.57\times$ for 8B, 14B, and 32B, respectively.

\begin{figure}[!t]
    \centering
    \begin{minipage}[t]{0.40\linewidth}
        \centering
        \includegraphics[width=\linewidth]{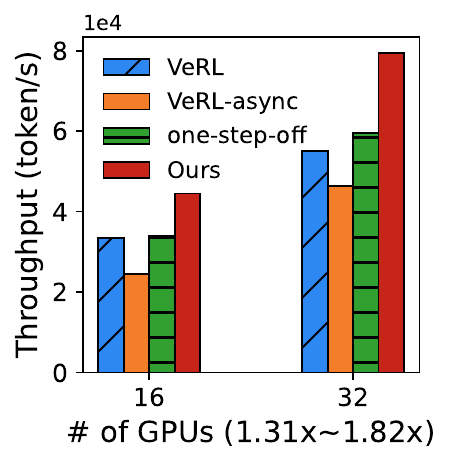}
        \caption*{(a)  Training throughput}
    \end{minipage}
    \hfill
    \begin{minipage}[t]{0.5\linewidth}
        \centering
        \includegraphics[width=\linewidth]{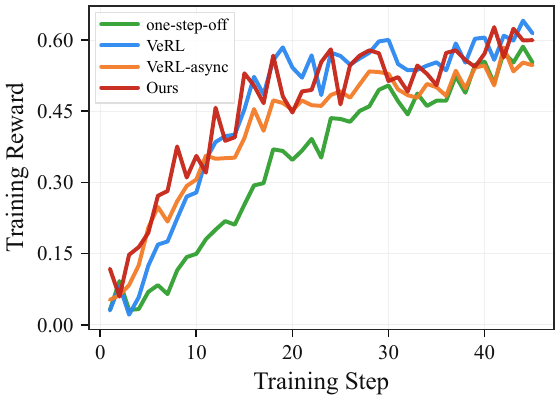}
        \caption*{(b) Training reward}
    \end{minipage}

    \caption{The training performance on Retool agentic tasks for Qwen3-8B. 
    }
    \label{fig:agent_performance}
    \vspace{-4mm}
\end{figure}

\noindent\textbf{Multi-Turn Agentic Tasks.}
These gains extend to workloads involving repeated generation and environment interaction.
For APPS coding tasks, Figure~\ref{fig:throughput}~(b) shows that \sysname{} outperforms all baselines for Qwen3-4B, 8B, and 14B on 8--64 GPUs, achieving speedups of $1.07\times$--$1.92\times$, $1.17\times$--$1.67\times$, and $1.36\times$--$1.71\times$, respectively.
For ReTool tool-interaction tasks, \sysname{} further achieves $1.31\times$--$1.82\times$ higher throughput with Qwen3-8B on 16 and 32 GPUs (Figure~\ref{fig:agent_performance}~(a)).
Together, these results demonstrate the effectiveness of \sysname{} across both single-turn and multi-turn RL workloads.

These improvements come from two complementary designs. The dual replay data cycle reduces fresh-rollout demand through trajectory reuse and mitigates trainer stalls caused by variable rollout completion times, while dynamic resource placement and adaptive scheduling reduce pipeline idle time under changing workloads. We further analyze their contributions in Sec.~\ref{subsec:performance_analysis}.

\subsection{Training Convergence}

We evaluate reward progression over both training steps and wall-clock time to assess whether \sysname{} accelerates training while preserving convergence.

\begin{figure}[!t]
    \centering
    \begin{minipage}[t]{0.48\linewidth}
        \centering
        \includegraphics[width=\linewidth]{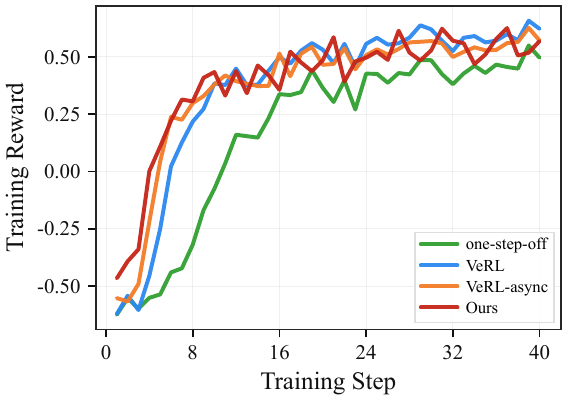}
        \caption*{(a)  Qwen3-8B on 32 GPUs}
    \end{minipage}
    \hfill
    \begin{minipage}[t]{0.48\linewidth}
        \centering
        \includegraphics[width=\linewidth]{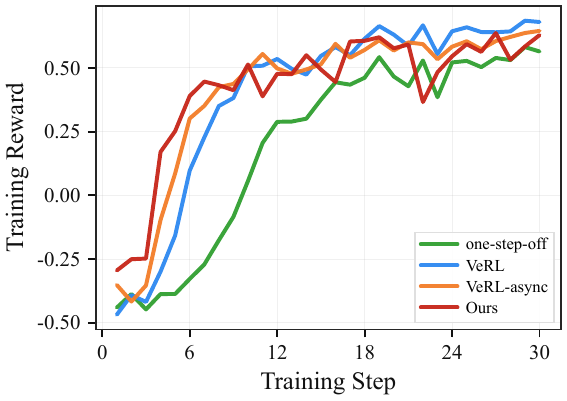}
        \caption*{(b) Qwen3-14B on 64 GPUs}
    \end{minipage}

    \caption{Training reward with respect to step.}
    \label{fig:training_reward}
\end{figure}

\begin{figure}[!t]
    \centering
    \begin{minipage}[t]{0.48\linewidth}
        \centering
        \includegraphics[width=\linewidth]{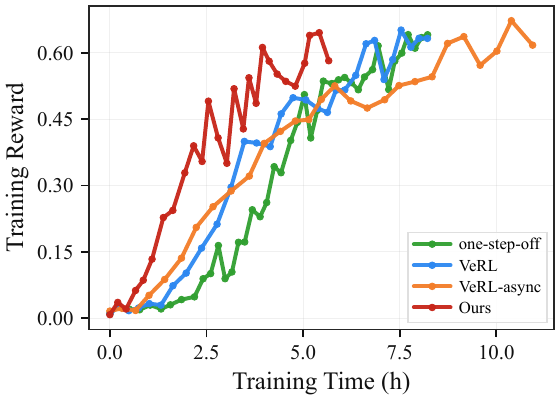}
        \caption*{(a)  Qwen3-4B on 8 GPUs}
    \end{minipage}
    \hfill
    \begin{minipage}[t]{0.48\linewidth}
        \centering
        \includegraphics[width=\linewidth]{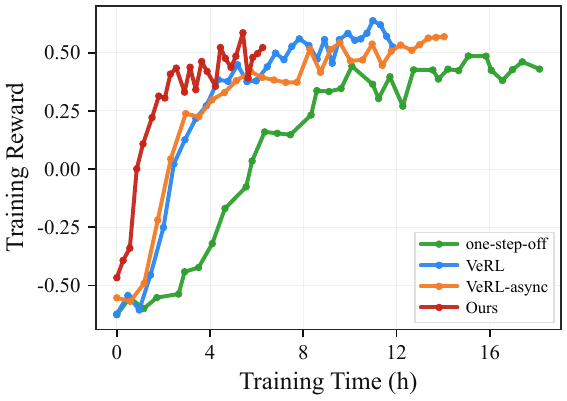}
        \caption*{(b) Qwen3-8B on 32 GPUs}
    \end{minipage}

    \caption{Training reward with respect to wall clock time.}
    \label{fig:training_reward_time}
    \vspace{-4mm}
\end{figure}

\noindent\textbf{Single-Turn Reasoning.}
Figure~\ref{fig:training_reward} shows that \sysname{} achieves final training rewards comparable to synchronous and asynchronous baselines for both Qwen3-8B and Qwen3-14B, while maintaining competitive early-stage progress.
These results demonstrate that trajectory reuse does not compromise convergence under our replay-driven dataflow.
Moreover, the Qwen3-8B results in Figure~\ref{fig:training_reward_time}~(b) show that \sysname{} reaches comparable reward levels substantially earlier, confirming that its system efficiency gains translate into shorter training time.

\noindent\textbf{Multi-Turn Agentic Tasks.}
The benefits also extend to multi-turn workloads with environment interactions.
For the Qwen3-4B coding agent, Figure~\ref{fig:training_reward_time}~(a) shows that \sysname{} improves reward substantially faster in wall-clock time, reaching reward levels comparable to the baselines within a shorter training period.
For ReTool, Figure~\ref{fig:agent_performance}~(b) shows that \sysname{} maintains reward progression comparable to VeRL and VeRL-async over training steps and consistently outperforms one-step-off throughout most of training.
Despite fluctuations during optimization, reward continues to improve without sustained degradation, supporting the stability of our design in multi-turn agentic RL.

Together, these results show that \sysname{} preserves stable learning across single-turn and multi-turn tasks while translating system throughput gains into faster time to target reward.

\vspace{-6mm}
\subsection{Performance Analysis}\label{subsec:performance_analysis}

\begin{figure}[!t]
    \centering
    \includegraphics[width=\linewidth]{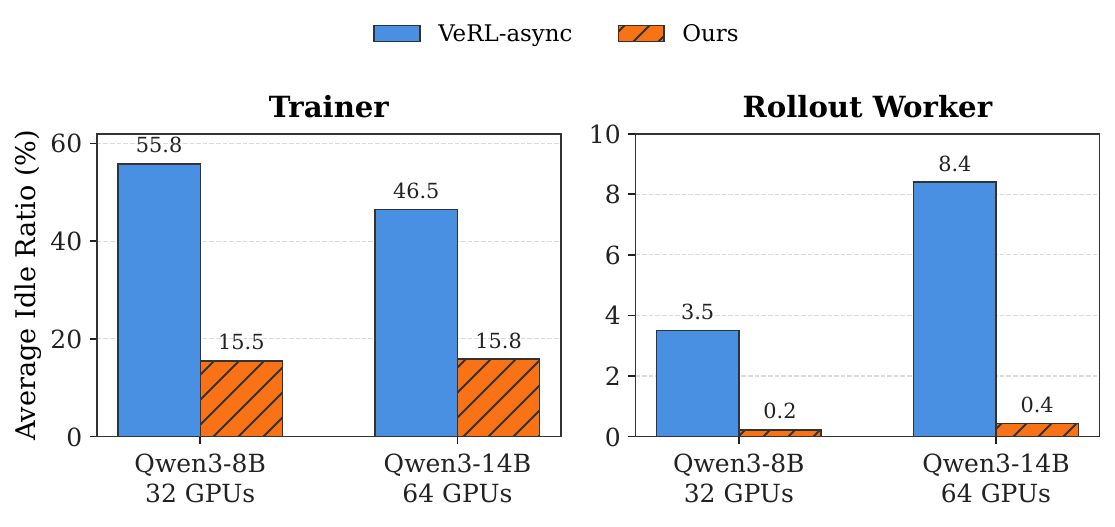}
    \caption{Trainer and rollout worker idle ratio comparison. }
    \label{fig:idle_performance}
    \vspace{-6mm}
\end{figure}

\noindent\textbf{Resource efficiency of \sysname{}.}
Figure~\ref{fig:idle_performance} compares the average idle ratios of trainers and rollout workers across two model scales.
We define the idle ratio as the fraction of each step time during which a worker remains idle, and report the average across all steps in the convergence experiments.
Compared with \textsc{VeRL-async}, \sysname{} reduces trainer idle ratios from 55.8\% to 15.5\% for Qwen3-8B on 32 GPUs and from 46.5\% to 15.8\% for Qwen3-14B on 64 GPUs.
Rollout worker idle ratios are further reduced from 3.5\% to 0.2\% and from 8.4\% to 0.4\%, respectively.

These results show that \sysname{} effectively reduces cross-stage waiting: replay-driven decoupling mitigates trainer stalls from data dependency, while keeping rollout workers nearly continuously utilized.
This improved resource utilization directly contributes to the throughput gains in Figure~\ref{fig:throughput}.

\noindent\textbf{Overhead of dynamic resource placement.}
Table~\ref{tab:drp_performance} reports the overhead of reallocating GPUs between rollout and training on a 32-GPU setup.
Reconfiguring from $16{:}16$ to $24{:}8$ and back to $16{:}16$ takes only 5.72s and 9.20s, respectively.
These results show that \sysname{} can adapt its rollout--training resource split at runtime with only lightweight overhead.
This is enabled by the Unified AsyncHybrid Worker design, which allows workers to switch roles without costly system-level reinitialization, full worker teardown, or long execution stalls.
As a result, \sysname{} can support adaptive resource placement during long-running RL workloads while keeping rebalancing overhead small relative to the overall training process.
\begin{table}[!t]
\centering
\caption{Dynamic resource placement overhead in \sysname{}.}
\label{tab:drp_performance}
\begin{tabular}{ccc}
\hline
Rollout : Train Setting & Direction & Cost (s)\\
\hline
32 GPUs (16:16 $\rightarrow$ 24:8) & train to rollout  &  5.72\\
32 GPUs (24:8 $\rightarrow$ 16:16)& rollout to train & 9.20 \\
\hline
\end{tabular}
\end{table}

\begin{table}[!t]
\centering
\caption{Effect of adaptive generation scheduling in \sysname{}.}
\label{tab:generation}
\begin{tabular}{cc}
\hline
Setting &  Average Generation\\
&  Throughput (token/s)\\
\hline
w/o adaptive scheduling & 1585.1 (+0.0\%) \\
w/ adaptive scheduling& 1749.8 (+10.4\%) \\
\hline
\end{tabular}
\vspace{-2mm}
\end{table}

\vspace{1mm}
\noindent\textbf{Efficiency of adaptive generation scheduling.}
Table~\ref{tab:generation} evaluates the impact of adaptive generation scheduling when training Qwen3-4B on 4 GPUs.
Compared with the case without generation scheduling, enabling the adaptive scheduling improves generation throughput from 1585.1 to 1749.8 tokens per second, achieving a 10.4\% gain.
This
reveals effectively improved runtime generation efficiency by better coordinating generation workloads, without introducing additional resource requirements.

\vspace{-2mm}
\section{Conclusion}\label{sec:conclusion}

We present \sysname{}, a fully decoupled and adaptive framework for efficient LLM RL post-training. 
Introducing a dual replay-driven cyclic dataflow, \sysname{} decouples rollout generation and training while enabling trajectory reuse, realizing co-optimization of system efficiency and training stability. 
Combined with efficient dynamic resource placement and adaptive resource scheduling, \sysname{} effectively improves system efficiency under dynamic and evolving RL workloads. 
Experimental results on up to 128 GPUs show that \sysname{} achieves up to 2.06$\times$ throughput gains while maintaining comparable or even better model performance.
By substantially reducing the cost of policy-improvements, \sysname{} opens up opportunities for more compute-intensive and iterative post-training paradigms, such as recursive self-improvement.
\section*{Availability}

We are currently organizing the code and awaiting the necessary approvals. The code will be released once the approval process is complete.

\newpage

\bibliographystyle{plain}
\bibliography{reference}


\end{document}